# Deep Learning Phase Segregation

Amir Barati Farimani,[1,][*] Joseph Gomes,[1,][*] Rishi Sharma,[2] Franklin L. Lee,[3] and Vijay S. Pande[1,][†]

[1]*Department of Chemistry, Stanford University, Stanford, California 94305*

[2]*Department of Electrical Engineering,*
*Stanford University, Stanford, California 94305*

[3]*Department of Chemical Engineering,*
*Stanford University, Stanford, California 94305*

(Dated: March 10, 2018)

## Abstract

Phase segregation, the process by which the components of a binary mixture spontaneously separate, is a key process in the evolution and design of many chemical, mechanical, and biological systems. In this work, we present a data-driven approach for the learning, modeling, and prediction of phase segregation. A direct mapping between an initially dispersed, immiscible binary fluid and the equilibrium concentration field is learned by conditional generative convolutional neural networks. Concentration field predictions by the deep learning model conserve phase fraction, correctly predict phase transition, and reproduce area, perimeter, and total free energy distributions up to 98% accuracy.

Phase segregation has many applications in chemical, biological, metallurgical, and mechanical systems. This phenomena plays a fundamental role in imparting mechanical, chemical, and electrical properties, among others, to materials. It has many practical consequences, ranging from cellular components separation and DNA sequencing,[1,2] to microstructural engineering in materials science[3] and the stabilization of colloidal suspensions.[4,5] For example, predicting and controlling the phase segregation and the phase domain size is of utmost significance in designing efficient organic solar cells.[6–9]

The predominant computational model for understanding the macroscopic time evolution of the phase segregation process by which an initially dispersed binary fluid spontaneously separates and forms domains pure in each component is approximated by the Cahn-Hilliard equation.[10–12] Generally, these equations are solved numerically, the downsides of which are two-fold: first, the equations are solved through a computationally intensive iterative scheme to predict the final decomposition; second, the process of decomposition is not reversible.

Recently, there has been a renaissance in the fields of computer vision[13] and natural language processing[14] due to advances in the methodology of deep learning.[15,16] The flexibility of deep neural networks allows models in principle to learn successively higher orders of features from the simplest possible representations of the data. Generative deep learning models, a class of unsupervised deep learning models, are capable of learning the important characteristics of a data distribution and generating realistic samples based purely on observation. Conditional generative models learn a direct mapping between a set of conditioning variables and the desired data distribution and have been used for artistic style transfer, text to image translation, and image to image translation.[17–20] The ability to learn features from raw data in an unsupervised fashion makes the application of deep learning models in physics highly attractive. Recently, we have shown that conditional generative convolutional neural networks can be used to learn, infer, and predict continuum-level physics.[21]

In this work, we present a data-driven approach for the learning, modeling, and prediction of phase segregation based on conditional generative adversarial network (cGAN) models. The cGAN model architecture and training procedure are shown in Figure 1. A direct, reversible mapping between an initially dispersed, immiscible binary fluid and the equilibrium concentration field across a range of initial binary concentrations is learned. In contrast to the iterative solution procedure, our framework can learn and infer the non-linear physical phenomenon of phase segregation without any knowledge of the underlying physical laws. Concentration field predictions by the deep learning model conserve phase fraction, correctly predict the phase transition, and reproduce area, perimeter, and total free energy distributions. The ability to reverse engineer phase segregation phenomena has the potential to open up new avenues for the rational design of specific microstructures given the initial mixture concentration and the mixing strength.

We report the results of a phase segregation model that learns a direct mapping to a steady-state concentration field given a randomly sampled dispersed binary mixture (Figure 1b). Details on the construction of train and test datasets are given in Methods section (section C). The evaluation of the trained phase separation model on ten representative samples from the test set is demonstrated in Figure 2. The results show a very close prediction of the concentration field compared to the ground truth across a wide range of mean concentrations and randomized initial conditions. The model maps the relaxed phase segregation system with a high spatial accuracy. In addition, the phase transition between single and binary phase regions is successfully captured. In the two-phase region ($0.2 < c < 0.8$), the deep learning model correctly predicts the dominant phase in each case, and reproduces phase micro-structure and spatial organization. In the single-phase regions ($c < 0.2$ or $c > 0.8$), the model again predicts the dominant phase, however, there are very minor occurrences of two-phase concentration regions not present in the ground truth solution.

The cGAN generator neural network learns a direct mapping between the initial binary mixture and the phase segregated concentration field through a hierarchical generative process. The activation layer outputs of the generator neural network shown in Figure 1b demonstrate how spinodal decomposition inference occurs. During the encoding process, regions of similar concentration are identified and aggregated by convolutional downsampling. During the decoding process, the aggregated regions of similar concentration form seeds by which single-phase enriched regions manifest. The cGAN discriminator neural network learns to classify samples as real or generated. The activation outputs of the discriminator neural network shown in Figure 1c demonstrate how input classification is performed. The raw values of the final layer of the discriminator can be interpreted as the probability that a corresponding patch of the input data comes from the ground truth data distribution. During model training, the discriminator function is optimized to maximize the probability of correct classification, and the generator function is optimized to both reconstruct ground truth data and minimize the probability the discriminator function cor-rectly classifies fake data; the weight of reconstruction loss relative to misclassification loss is controlled by a hyperparameter ($\lambda$). We show empirically that reconstruction loss (L1) alone fails to capture high-frequency details of phase segregated liquid configurations (see *Supporting Information*).

To investigate the thermodynamic properties of samples generated by the cGAN model, we post-processed the outputs from deep learning models and compared them to the results of ground truth calculations. The summary of mean absolute and relative average errors for different hyperparameter values of $\lambda$ are tabulated in Table I. The best performance in terms of total energy error is achieved for the case of $\lambda = 1.0$ (i.e., the case of L1+1.0GAN). The detailed comparison between size and energy metrics for the case of L1+1.0GAN is depicted in Figure 4 for the test set. The computational algorithm used to compute the chemical free energy component, surface free energy component, and total free energy, along with mean concentration, phase area, and phase perimeter are elaborated in *Supporting Information*. The computed chemical, surface, and total Ginzburg-Landau free energies show a very good match between the model output and the ground truth (Figures 4c, 4d, and 4e, respectively). Here, the model's successful prediction of the phase transition (single phase to two phases) is demonstrated. The mean concentration of the predicted segregated phases and the initial mean mixed concentration match very well over the test set (Figure 4f). In addition, we compared the phase perimeter and area generated by the model and the ground truth for $\lambda = 1.0$ (Figures 4a and 4b, respectively) to demonstrate the preservation of domain boundaries.

Additionally, we report the results of a reversible phase segregation model, which maps from the segregated phase to the initial mixture (Figure 3). We expect such models could aid in the rational engineering and design of microstructured materials. The model can successfully map the concentration field while preserving the mean concentration. The mean concentration has a mean absolute error of 0.0029, while the standard deviation of the concentration field has a mean absolute error of 0.0021. The latter point emphasizes that the degree of fluctuation has been properly captured by the learning model. These are plotted in *Supporting Information*.

We have demonstrated that conditional generative deep learning models can be used to directly learn the physics of phase segregation based only on observations with high fidelity. We demonstrated successful learning and prediction for steady state spatial decomposition of the binary mixture, given a noisy initial concentration field. We also showed the ability to learn the reverse mapping between the phase segregated concentration field to initial binary mixture, useful for interface design and engineering. We observed that adversarial training of the generative model improved both geometric and thermodynamic validation metrics. Because phase segregation is a highly non-linear phenomena, we conceived that our framework is capable of generalizing to learn most non-linear physical phenomena.

## METHODS

### A. Conditional Generative Adversarial Networks

We adapt the conditional Generative Adversarial Network (cGAN), which was used previously for image-to-image translation.[22] cGANs are generative models that learn a mapping from observed data $c$ to output data $\hat{x}$: $G$: $c \to \hat{x}$. Here, $c$ is a representation of initially dispersed domain and $\hat{x}$ is the observed solution. The generator $G(c)$ is optimized to produce outputs $\hat{x}$ that cannot be distinguished from training data by a discriminator, $D$. $D(c, \hat{x})$ is a scalar that represents the probability that $\hat{x}$ came from $p_{model}(\hat{x})$ (the data distribution) rather than the output of $G(c)$. The generator $G$ and discrimator $D$ models are convolutional neural networks, adapted from Ref. [23]. The generator $G$ uses a "U-Net"-based network architecture[24] and the discriminator $D$ uses a convolutional "PatchGAN" classifier architecture.[25] The combined network architecture and training procedure are diagrammed in Figure 1a.

We train $D$ to maximize the probability of assigning the correct label to both training examples and samples from $G$. The cGAN objective function is expressed as:

$$\mathcal{L}_{cGAN}(G, D) = \mathbb{E}_{c,\hat{x} \sim p_{model}(c,\hat{x})}[log D(c, \hat{x})] + \\ \mathbb{E}_{c,\hat{x} \sim p_{model}(c,\hat{x})}[log(1 - D(c, G(c)))]. \quad (1)$$

Here, $D$ and $G$ participate in a two-player minimax game with value function $\mathcal{L}_{cGAN}(G, D)$, where $G$ attempts to minimize this objective against an adversarial $D$ that tries to maximize it, $G^* = arg\ min_G\ max_D\ \mathcal{L}_{cGAN}$. In addition, we apply an L1 distance loss function to the generator:

$$\mathcal{L}_{L1}(G) = \mathbb{E}_{c,\hat{x} \sim p_{model}(c,\hat{x})}[\|\hat{x} - G(c)\|_1]. \quad (2)$$

The final objective is:

$$G^* = \underset{G}{\operatorname{argmin}} \max_{D} \; \mathcal{L}_{L1}(G) + \lambda \mathcal{L}_{cGAN}(G, D) \qquad (3)$$

where the hyperparameter $\lambda$ is the cGAN weight.

The discriminator $D$ (Figure 1c) is a convolutional neural network that operates on either $(c, \hat{x})$, the input and generated output, or $(c, \hat{x}')$, the input and ground truth solution, to produce the probability that a small patch of the discriminator input comes from the training data distribution. In this way, the discriminator treats each real or generated sample as a Markov random field (MRF), an undirected probabilistic graph which assumes statistical independence between nodes separated by more than a patch diameter.[26–28] This operation is performed convolutionally across the entire solution, averaging all responses of all distinct patches to provide the output of $D$. By training the discriminator to correctly distinguish between real and generated samples (Methods, Equation 2), we build a joint probabilistic model over the desired field values at discrete grid points. In the image modeling community, the treatment of images as MRFs has been previously explored, and is commonly used in models to determine texture or style loss.[29,30]

For generator $G$ (Figure 1b), we adopt an encoder-decoder network, which has been used successfully in image and text translation.[31–33] The input representing the dispersed and well mixed binary mixture is processed by the encoder network (Figure 1b, top), a series of convolutional operations that progressively downsample the grid until a reduced latent representation is reached (Figure 1d, top). The process is then reversed by the decoder network (Figure 1b, bottom), with a series of convolutional operations that progressively upsample the reduced representation, directly generating the inferred solution (Figure 1d, bottom). Due to the great deal of low-level information shared between the input and output grids, we share information directly between equivalent size encoder and decoder convolutional layers by the use of skip connections.[24] Generation proceeds by sampling the conditional probability density for the state of each grid point given the known states of its neighboring data points. By training a loss function that rewards the generator $G$ for successfully confusing the discriminator $D$, in addition to reproducing the ground truth solution for known observations.

## B. Optimizing Contribution of GAN

When training the network described in A, the hyperparameter $\lambda$ determines the relative weight applied to the adversarial component in the loss function in equation 3. Table I summarizes the results of varying $\lambda$ from 0 to 100. The $L1$ component of the loss function ensures that the output approximates the ground truth images in an $L1$ sense, which gives the overall structure of the final solution. However, $L1$ loss alone fails to capture many of the subtler, high-frequency details of the phase segregated liquids (*see Supporting Information*). This can be seen in Table I by the improved performance in a number of metrics as we begin to add in a GAN component to the loss. This component of the loss is best conceived as a *learned component* of the loss, which is learned via the adversarial game played by generator and discriminator. The learned component of the loss is used to determine whether the final output images are realistic - as determined by the discriminator that is encouraged to learn the difference between real and fakes. In practice, this adds a sharpness to the images produced, whereas the $L1$ loss alone results in washed out, blurry images that look less realistic. By including an GAN component to the loss, however, the model occasionally hallucinates structure where the correct phase segregated solution has none.

Ultimately, we see a slight improvement by adding a learned component to our loss, but if we increase the weighting on this component too much relative to ground truth $L1$ prediction, our model starts to degrade in quality.

## C. Datasets

We consider a two-dimensional domain with no input or output influx. We wish to obtain a long-time (steady-state) solution of the concentration field, $c(x, y, t)$. The Cahn-Hilliard equation involves fourth-order spatial partial-differential operators:

$$\frac{\partial c}{\partial t} = D\nabla^2(f'(c) - \epsilon^2 \nabla^2 c) \quad (4)$$

The phase field is in the form of $c(x, y, t)$. We assume a constant diffusion coefficient ($D$). The $\epsilon$ parameter is the energetic penalty on gradients in the concentration field, which drives the phase segregation process and influences the mean size of single-phase regions. $f(c)$ is the free energy of the system that is generally selected in the form $f(c) = ac^2(c-1)^2$, which is the double well potential with the critical quench. The parameter $a$ defines the depth of the wells and we considered $a = 1.0$. We solved the equation for the long-time behavior $(t \to \infty)$.

We solved Equation 4 by semi-implicit Fourier-spectral method with periodic boundary conditions. The numerical integration scheme is elaborated in *Supporting Information*.

A dataset containing 8640 training samples was generated by numerical simulation, varying the initial concentration for [0.05-0.94] with the step size of 0.01 and initial amplitude of thermal noise constant at 0.1. At each concentration, 96 random initial conditions were generated. The domain size is a grid of 64x64. The training data consists of pairs of 1-channel, 64x64 grids; the first grid represents the initial mixed state (Figure 2, left) and the second grid contains the solved phase field Figure 2, center). Another 8640 test samples were generated by randomizing the initial condition and with a different concentrations compared to the training set.

### ACKNOWLEDGEMENTS

The Pande Group is broadly supported by grants from the NIH (R01 GM062868 and U19 AI109662) as well as gift funds and contributions from Folding@home donors. We acknowledge the generous support of Dr. Anders G. Frøseth and Mr. Christian Sundt for our work on machine learning.

### AUTHOR CONTRIBUTIONS STATEMENT

ABF and JG contributed equally to this work. RS optimized the network architecture and hyperparameters and performed training and tests. FLL post-processed the model outputs. VSP supervised the project.

### COMPETING FINANCIAL INTERESTS

VSP is a consultant and SAB member of Schrodinger, LLC and Globavir, sits on the Board of Directors of Apeel Inc, Freenome Inc, Omada Health, Patient Ping, Rigetti Computing, and is a General Partner at Andreessen Horowitz.

# SUPPORTING INFORMATION

The evaluation of train and test set statistical accuracy are given in Supporting Information. The training data and model codes will be available upon request and as soon as paper is published.

TABLE I. Mean absolute error of properties by model for the 8k dataset. Relative errors (in %) are shown in parentheses.

| Property | | L1+0GAN | L1+1GAN | L1+10GAN | L1+100GAN |
|---|---|---|---|---|---|
| Size | Perimeter | 25.1 (6.6) | 23.8 (6.3) | 35.5 (9.3) | 46.0 (12.1) |
| Metrics[a] | Area | 85.7 (2.1) | 90.4 (2.2) | 181.1 (4.4) | 255.5 (6.2) |
| Free | Chemical | 0.120 (10.4) | 0.098 (8.5) | 0.212 (18.5) | 0.199 (17.4) |
| Energy[b] | Surface | 1.75 (7.7) | 1.47 (6.5) | 3.56 (15.8) | 3.94 (17.5) |
| | Total | 1.67 (7.2) | 1.43 (6.2) | 3.40 (14.7) | 3.76 (16.2) |
| Concentration | | 0.007 (0.76) | 0.014 (1.48) | 0.015 (1.62) | 0.012 (1.25) |

[a] in pixels.    [b] in kJ/mol.

[31] G. E. Hinton and R. R. Salakhutdinov, science **313**, 504 (2006).

[32] X. Wang and A. Gupta, in *European Conference on Computer Vision* (Springer, 2016) pp. 318–335.

[33] D. Yoo, N. Kim, S. Park, A. S. Paek, and I. S. Kweon, in *European Conference on Computer Vision* (Springer, 2016) pp. 517–532.
11

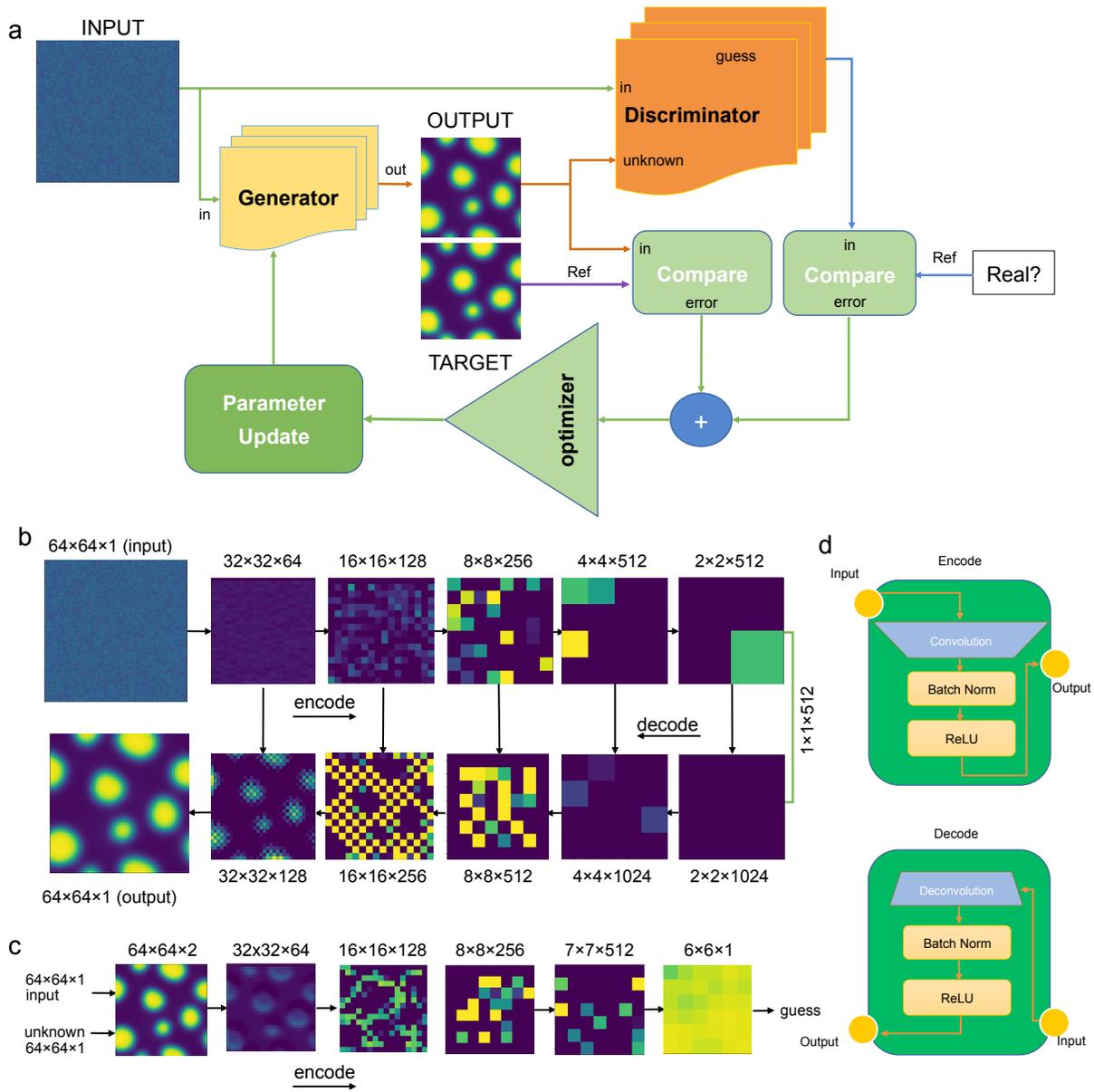

FIG. 1. **Conditional Generative Adversarial Network (cGAN) architecture. a.** A flowchart of cGAN model and the connections between input (binary phase mixture), output (concentration field), and model optimization procedure. **b.** The architecture of the convolutional neural network used for the cGAN generator. A representative output for each layer of the generator is shown. **c.** The architecture of the convolutional neural network used for the cGAN discriminator. A representative output for each layer of the discriminator is shown for ground truth sample. **d.** The architecture of the encode and decode modules used in the cGAN model.

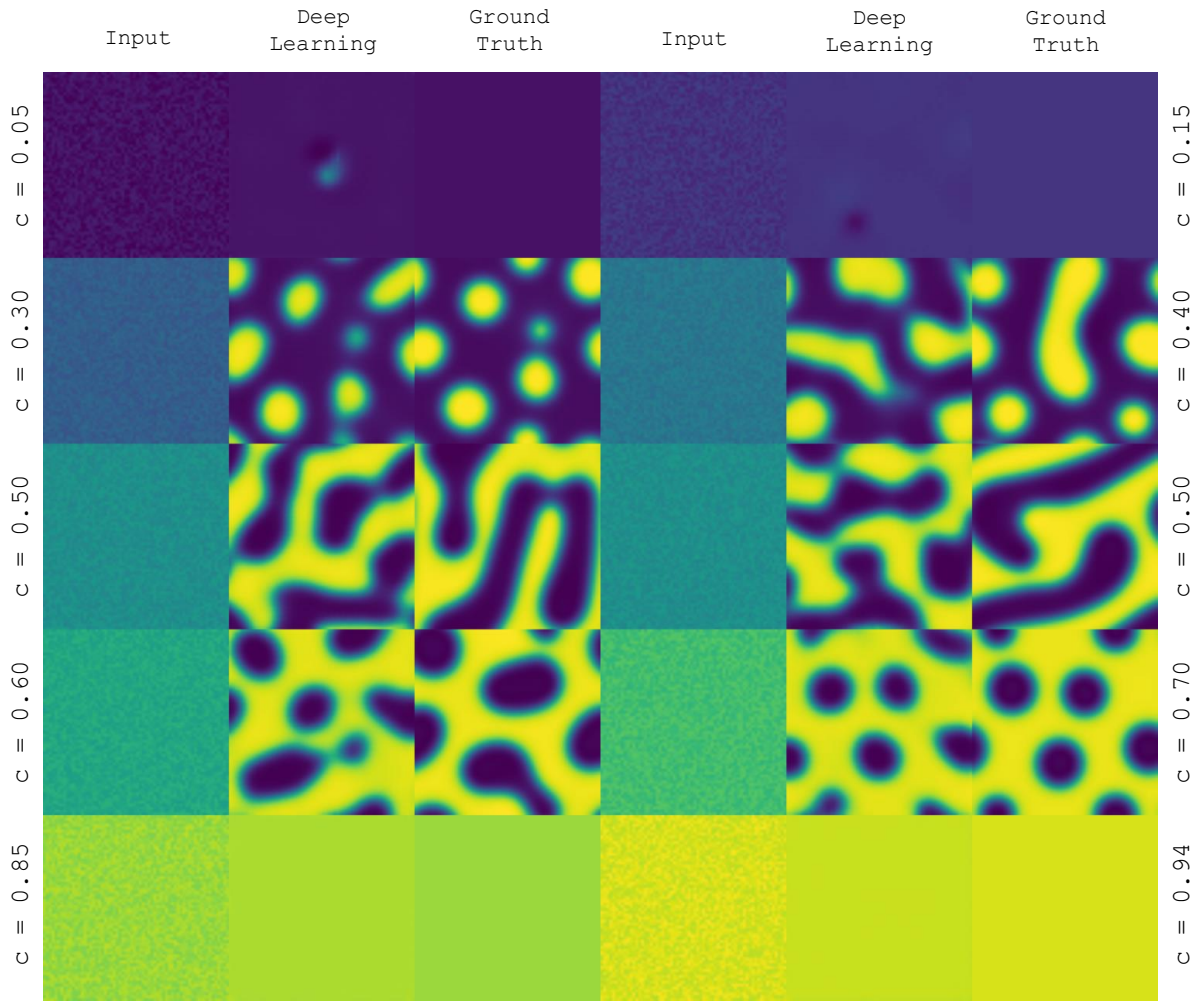

FIG. 2. **L1+1.0GAN performance on the phase separation 8k test set.** Long-time phase separation behavior of a binary mixture predicted by deep learning. The initial input distribution of mixture (left), the predicted solution (center), the ground truth solution of the phase field (right) are shown for the concentration range c = 0.05-0.94. There are two samples represented for c = 0.50 to demonstrate the variety of outcomes from different initial conditions with the same mean concentration.

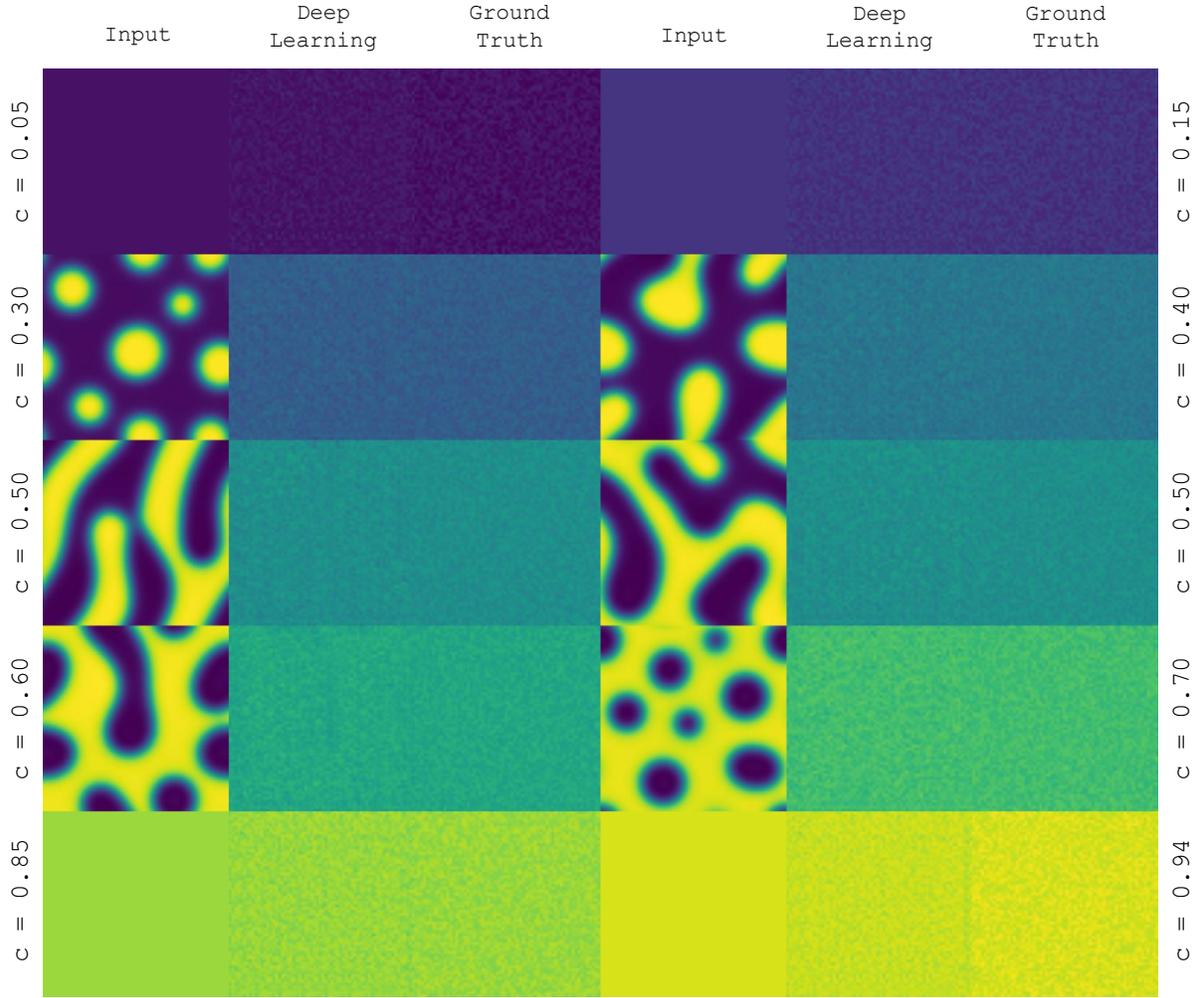

FIG. 3. **cGAN performance on the reverse phase separation test.** Long-time reverse phase separation behavior of a binary mixture predicted by deep learning. The input distribution of a phase separated mixture (left), the predicted initial phase field (center), the ground truth initial phase field (right) are shown for the concentration range c = 0.05-0.94. There are two samples represented for c = 0.50 to demonstrate the variety of outcomes.

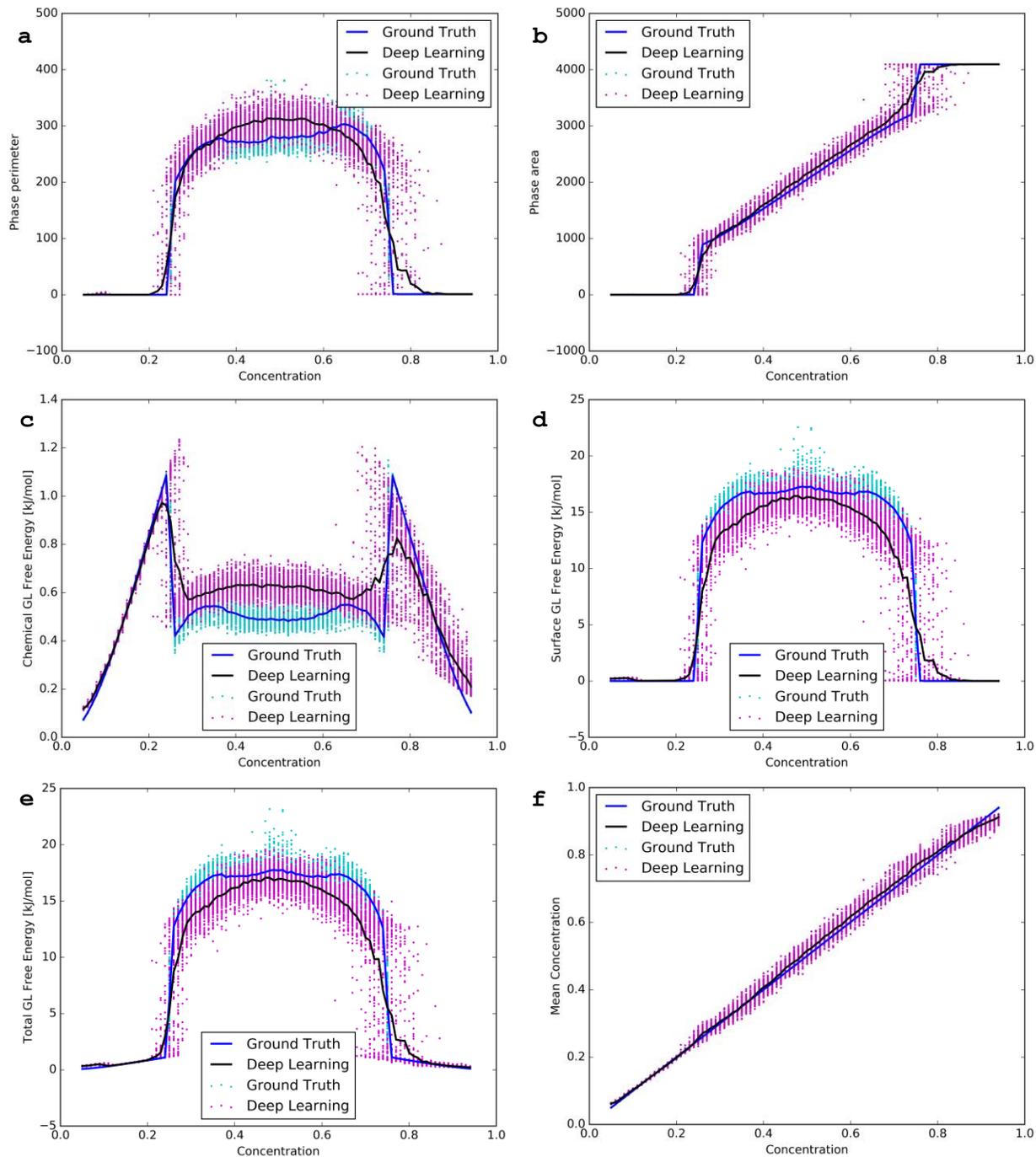

FIG. 4. **Comparison of metrics between deep learning model and ground truth for the L1+1.0GAN model on the 8k test set. a.** Phase A perimeter. **b.** Phase A area. **c.** Chemical free energy. **d.** Surface free energy. **e.** Total free energy. **f.** Mean phase A concentration. The x-axis in all cases represents the initial mean concentration of phase A.

# Supporting Information for:
# Deep Learning Phase Segregation

# I. SOLUTION TO THE CAHN-HILLIARD EQUATION

Phase separation has many applications in chemical, biological, metallurgical and mechanical systems. This phenomena is studied using the Cahn-Hilliard equation. The Ginzburg-Landau free energy is used to derive the Cahn-Hilliard equation and represents the sum of the surface and chemical free energies over the whole domain[1,2]:

$$\mathscr{E}(c) = \int_\Omega \Psi_c + \Psi_s d\boldsymbol{x} \tag{1}$$

The main quantity of interest is the phase field, which is $c(x, y, t)$. The chemical free energy $\Psi_c$ is then expressed as:

$$\Psi_c = f(c) \tag{2}$$

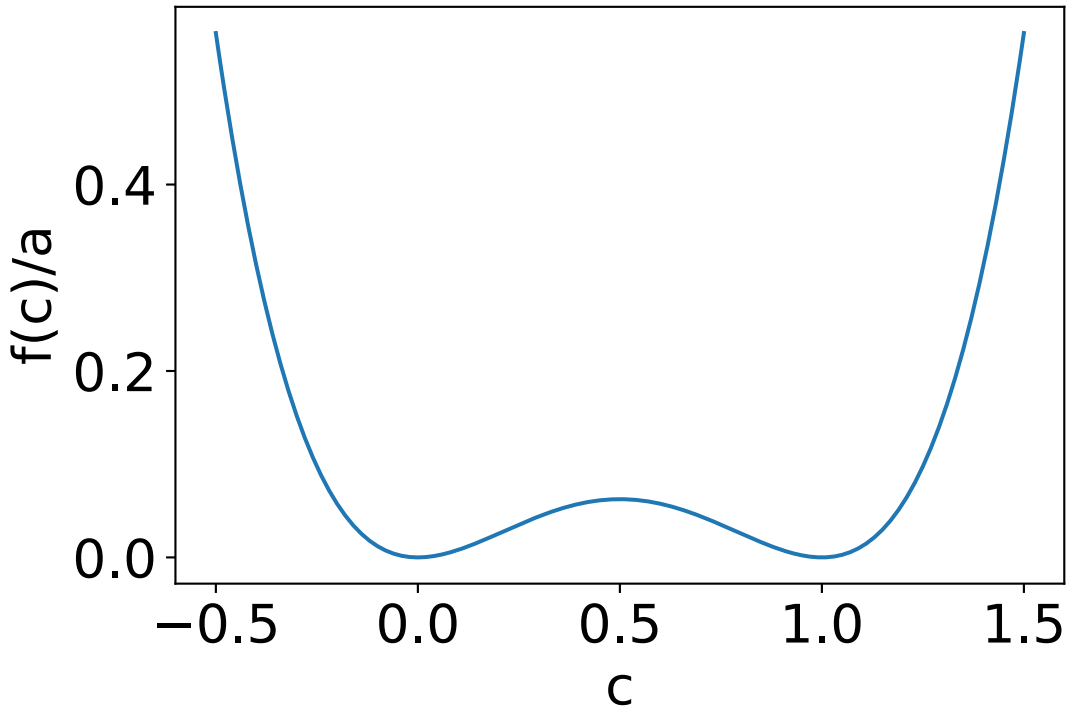

FIG. S1. *Double well potential chosen to represent the free energy of the system.*

Here, we selected the form $f(c) = ac^2(c-1)^2$, which is the double well potential with the critical quench (Figure S1). The parameter $a$ defines the depth of the wells; here, we selected $a = 1.0$. Meanwhile, the surface free energy $\Psi_s$ is expressed as:

$$\Psi_s = \frac{\epsilon^2}{2}|\nabla c|^2 \tag{3}$$

The chemical potential $\mu$ is then expressed as the derivative of $\mathscr{E}$ with respect to concentration, $\mu = f'(c) - \epsilon^2 \nabla c$. The net flux is defined as $J = -D\nabla\mu$, where $D$ is a positive constant diffusion constant. The continuity equation $\dfrac{\partial c}{\partial t} = -\nabla \cdot J$ is enforced, which leads to the Cahn-Hilliard equation:

$$\frac{\partial c}{\partial t} = D\nabla^2(f'(c) - \epsilon^2 \nabla^2 c) \tag{4}$$

To numerically solve the equation, it is easier to recast the equation in Fourier space and then propagate $\tilde{\phi}(\boldsymbol{k}, t)$ over time, where we define a transformed variable $\phi = 2c - 1$. Equation 4 then becomes:

$$\frac{\partial \tilde{\phi}}{\partial t} = -D|\boldsymbol{k}|^2(\tilde{f}'(\phi) - \epsilon^2 |\boldsymbol{k}|^2 \tilde{\phi}) \tag{5}$$

To propagate the phase field, Equation 5 is discretized in time:

$$\frac{\tilde{\phi}(\boldsymbol{k}, t+dt) - \tilde{\phi}(\boldsymbol{k}, t)}{dt} = -D|\boldsymbol{k}|^2(\tilde{f}'(\phi) - \epsilon^2 |\boldsymbol{k}|^2 \tilde{\phi}(\boldsymbol{k}, t+dt)) \tag{6}$$

Finally, rearranging for $\tilde{\phi}(\boldsymbol{k}, t+dt)$:

$$\tilde{\phi}(\boldsymbol{k}, t+dt) = \frac{\tilde{\phi}(\boldsymbol{k}, t) - D|\boldsymbol{k}|^2 \tilde{f}'(\phi)dt}{1 + \epsilon^2 |\boldsymbol{k}|^4 Ddt} \tag{7}$$

## II. EVALUATION OF PHASE SEPARATION MODELS

The evaluation of the trained phase separation models on ten representative samples from each of the training sets of the the L1+0.0GAN, L1+1.0GAN, L1+10.0GAN, L1+100.0GAN is demonstrated in Figures S2, S3, S4, and S5, respectively. The results show good reproduction of the concentration field compared to ground truth. The input concentration distributions are shown in the first and fourth columns of the figures. The ground truth distributions are shown in the second and fifth columns and the predicted distributions are shown in the third and sixth columns. Evaluation of samples from the test sets of the L1+0.0GAN, L1+10.0GAN, L1+100.0GAN models is demonstrated in Figures S6, S7, and S8, respectively.

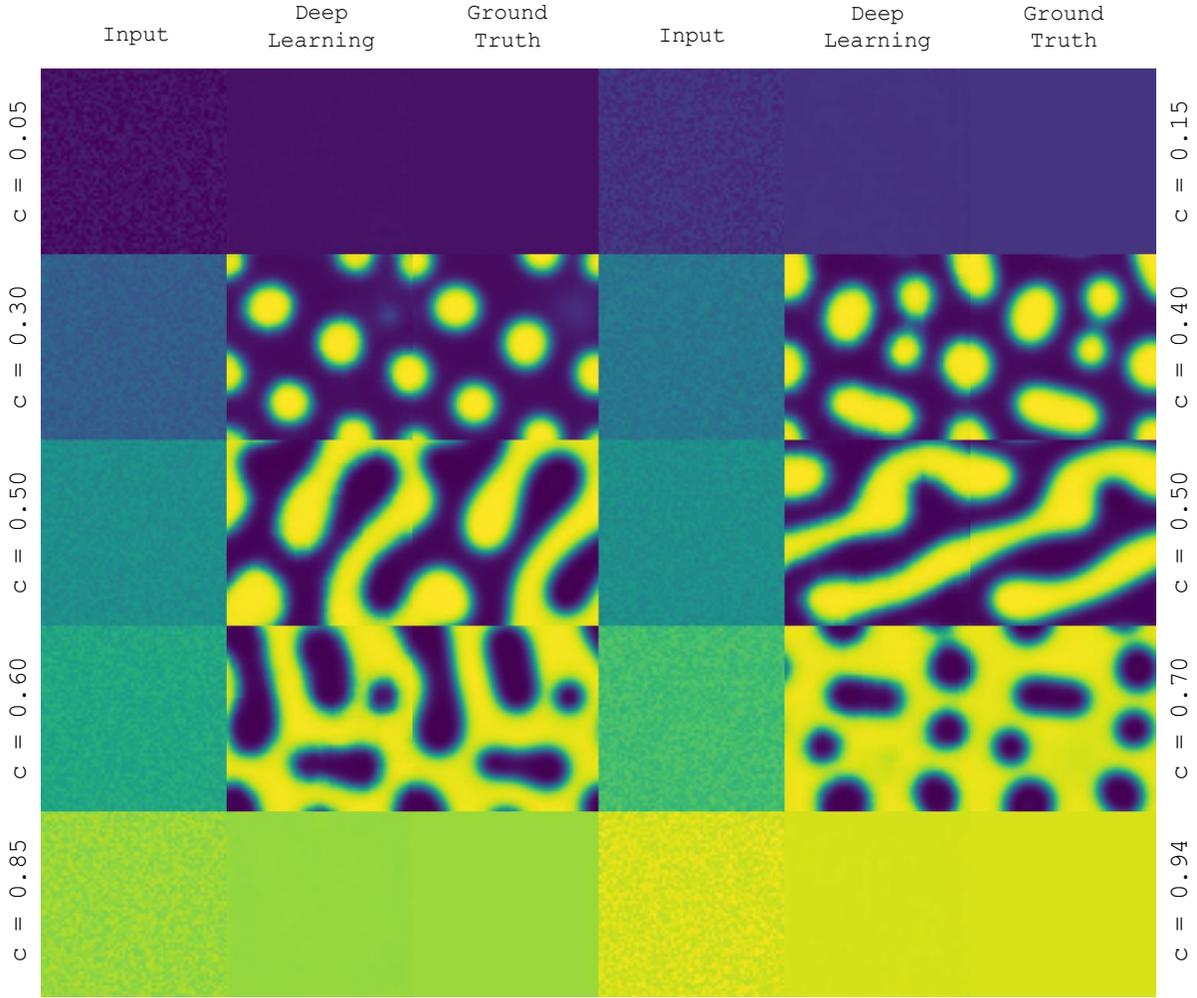

FIG. S2. **L1+0.0GAN performance on the phase segregation training set.** Long-time phase segregation behavior of a binary mixture predicted by deep learning. The initial input distribution of mixture (left), the predicted solution (center), the ground truth solution of the phase field (right) are shown for the concentration range c = 0.05-0.94. There are two samples represented for c = 0.50 to demonstrate the variety of outcomes.

## III. CALCULATION OF COMPARATIVE METRICS

The average concentration over the whole domain is calculated and compared to the initial value. Additionally for the reverse phase segregation case, the standard deviation of the concentration field is calculated to demonstrate the consistency of the underlying fluctuations between the target and the output (Figure S9).

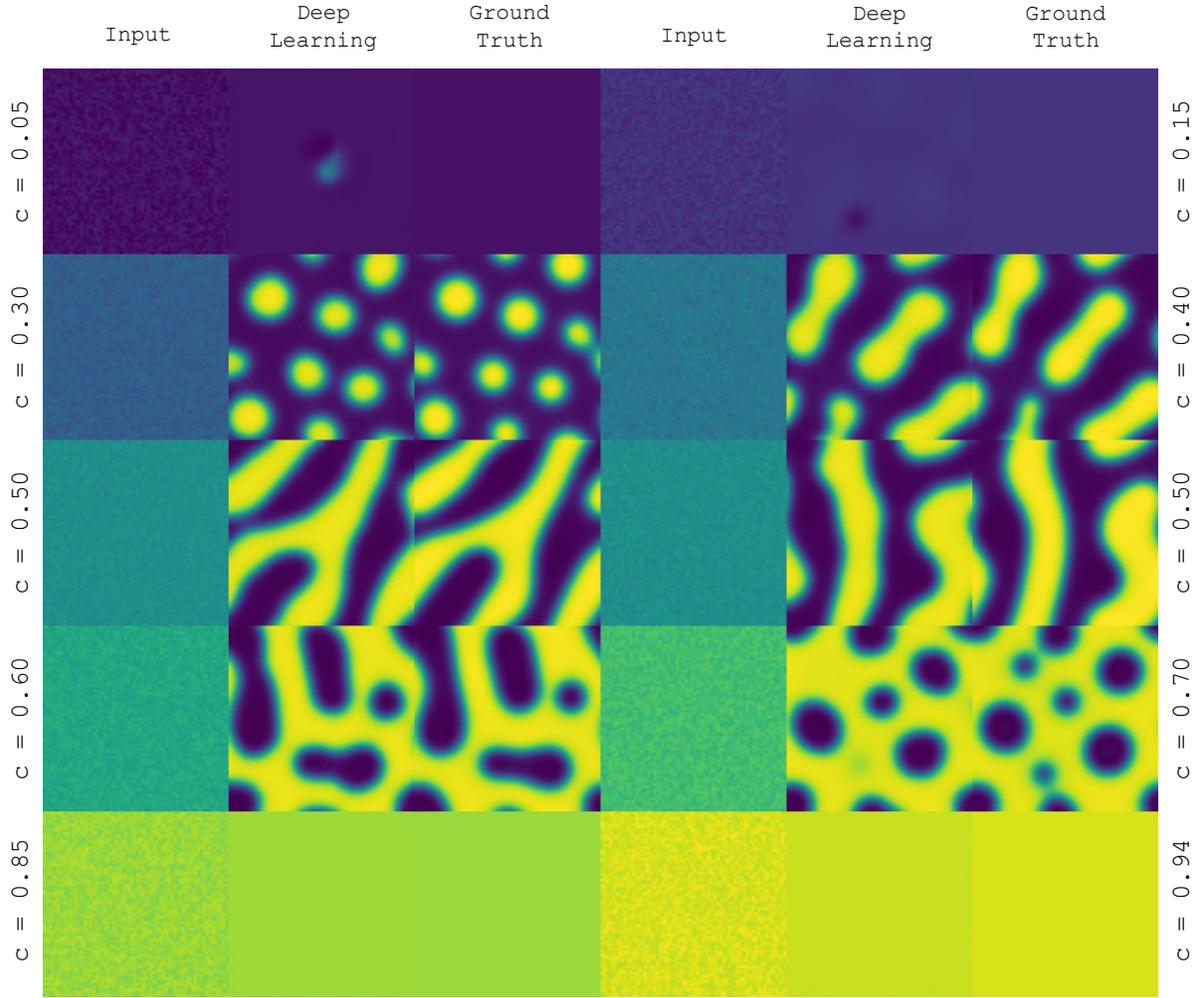

FIG. S3. **L1+1.0GAN performance on the phase segregation training set.** Long-time phase segregation behavior of a binary mixture predicted by deep learning. The initial input distribution of mixture (left), the predicted solution (center), the ground truth solution of the phase field (right) are shown for the concentration range c = 0.05-0.94. There are two samples represented for c = 0.50 to demonstrate the variety of outcomes.

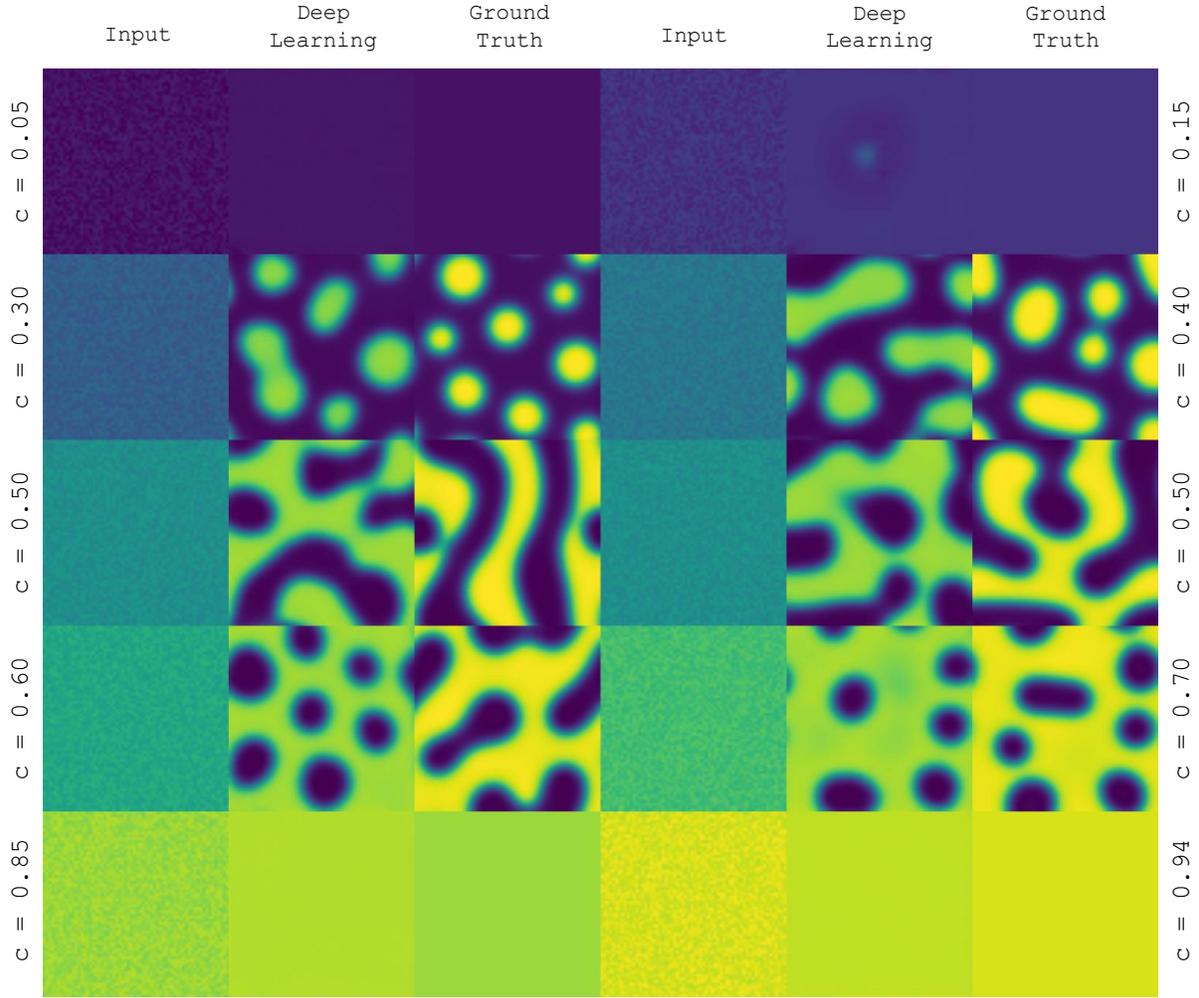

FIG. S4. **L1+10.0GAN performance on the phase segregation training set.** Long-time phase segregation behavior of a binary mixture predicted by deep learning. The initial input distribution of mixture (left), the predicted solution (center), the ground truth solution of the phase field (right) are shown for the concentration range c = 0.05-0.94. There are two samples represented for c = 0.50 to demonstrate the variety of outcomes.

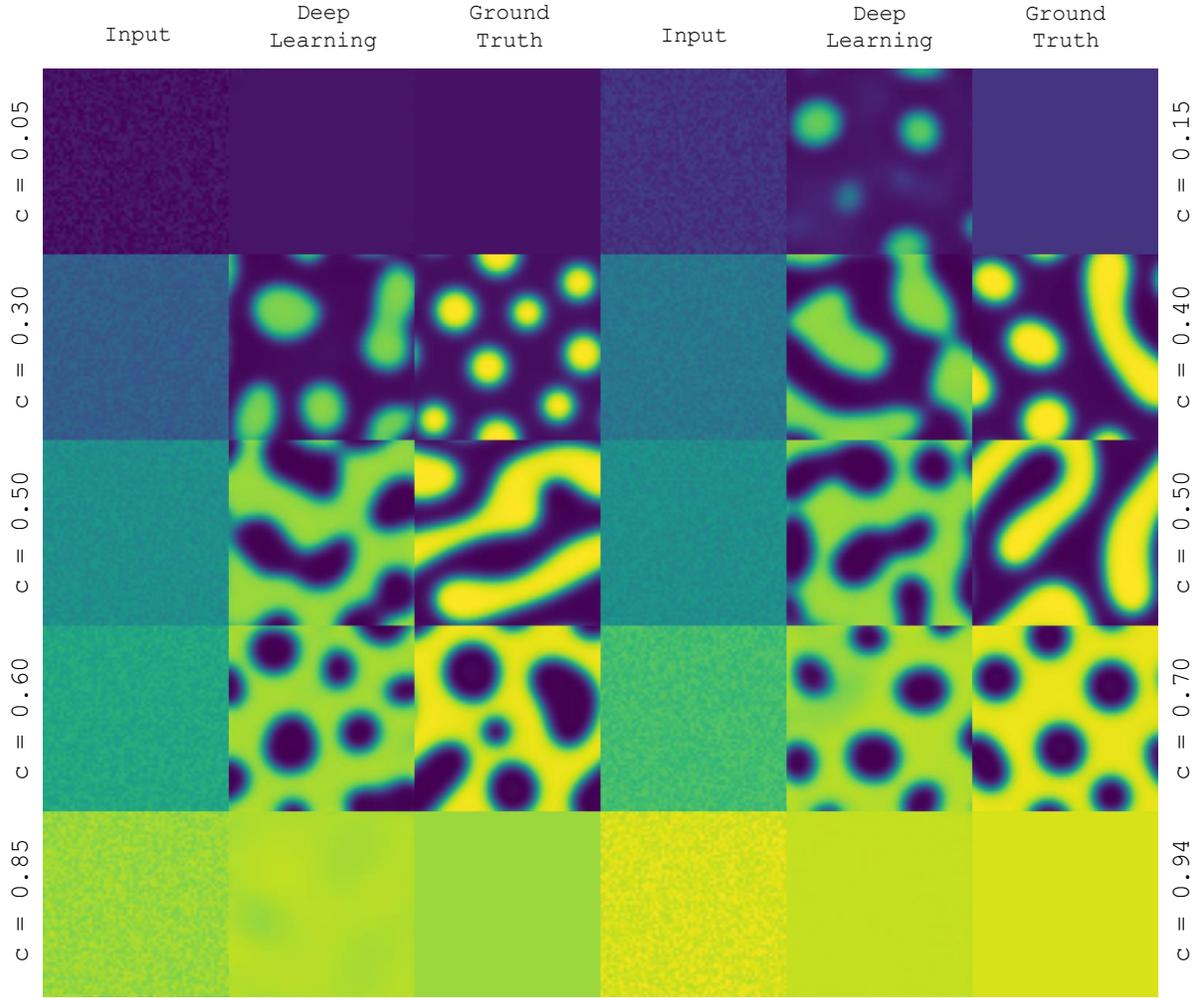

FIG. S5. **L1+100.0GAN performance on the phase segregation training set.** Long-time phase segregation behavior of a binary mixture predicted by deep learning. The initial input distribution of mixture (left), the predicted solution (center), the ground truth solution of the phase field (right) are shown for the concentration range c = 0.05-0.94. There are two samples represented for c = 0.50 to demonstrate the variety of outcomes.

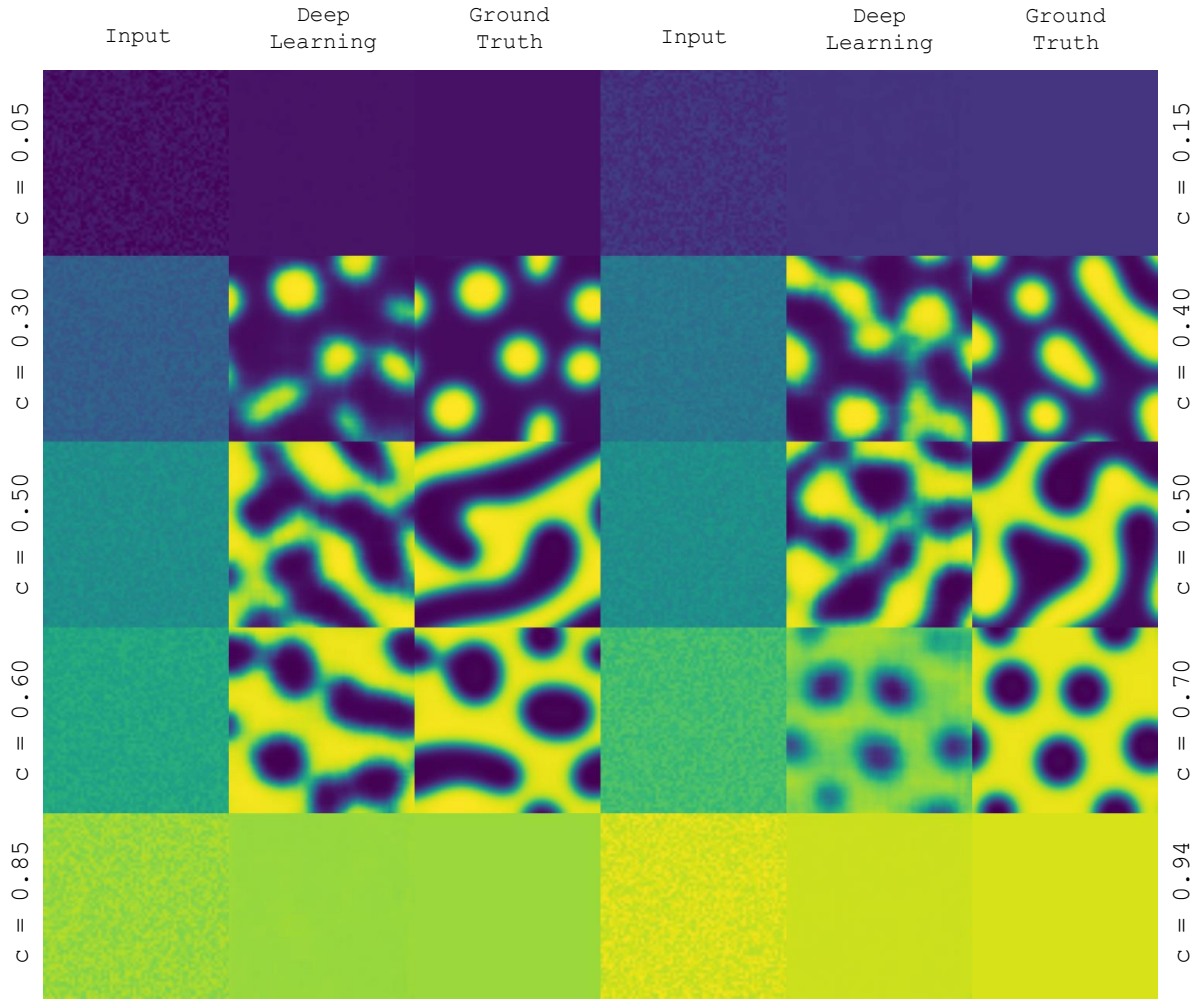

FIG. S6. **L1+0.0GAN performance on the phase segregation test set.** Long-time phase segregation behavior of a binary mixture predicted by deep learning. The initial input distribution of mixture (left), the predicted solution (center), the ground truth solution of the phase field (right) are shown for the concentration range c = 0.05-0.94. There are two samples represented for c = 0.50 to demonstrate the variety of outcomes.

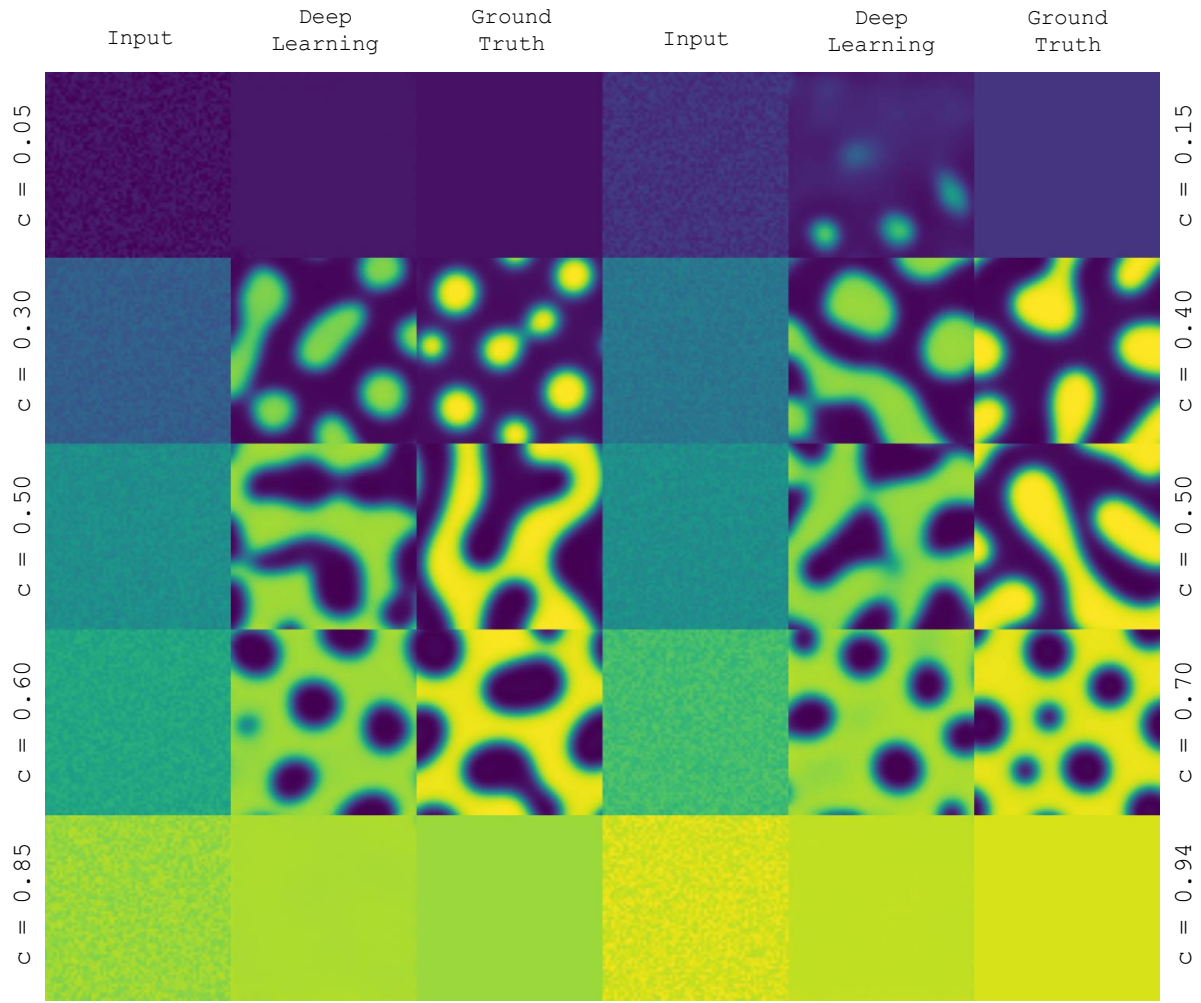

FIG. S7. **L1+10.0GAN performance on the phase segregation test set.** Long-time phase segregation behavior of a binary mixture predicted by deep learning. The initial input distribution of mixture (left), the predicted solution (center), the ground truth solution of the phase field (right) are shown for the concentration range c = 0.05-0.94. There are two samples represented for c = 0.50 to demonstrate the variety of outcomes.

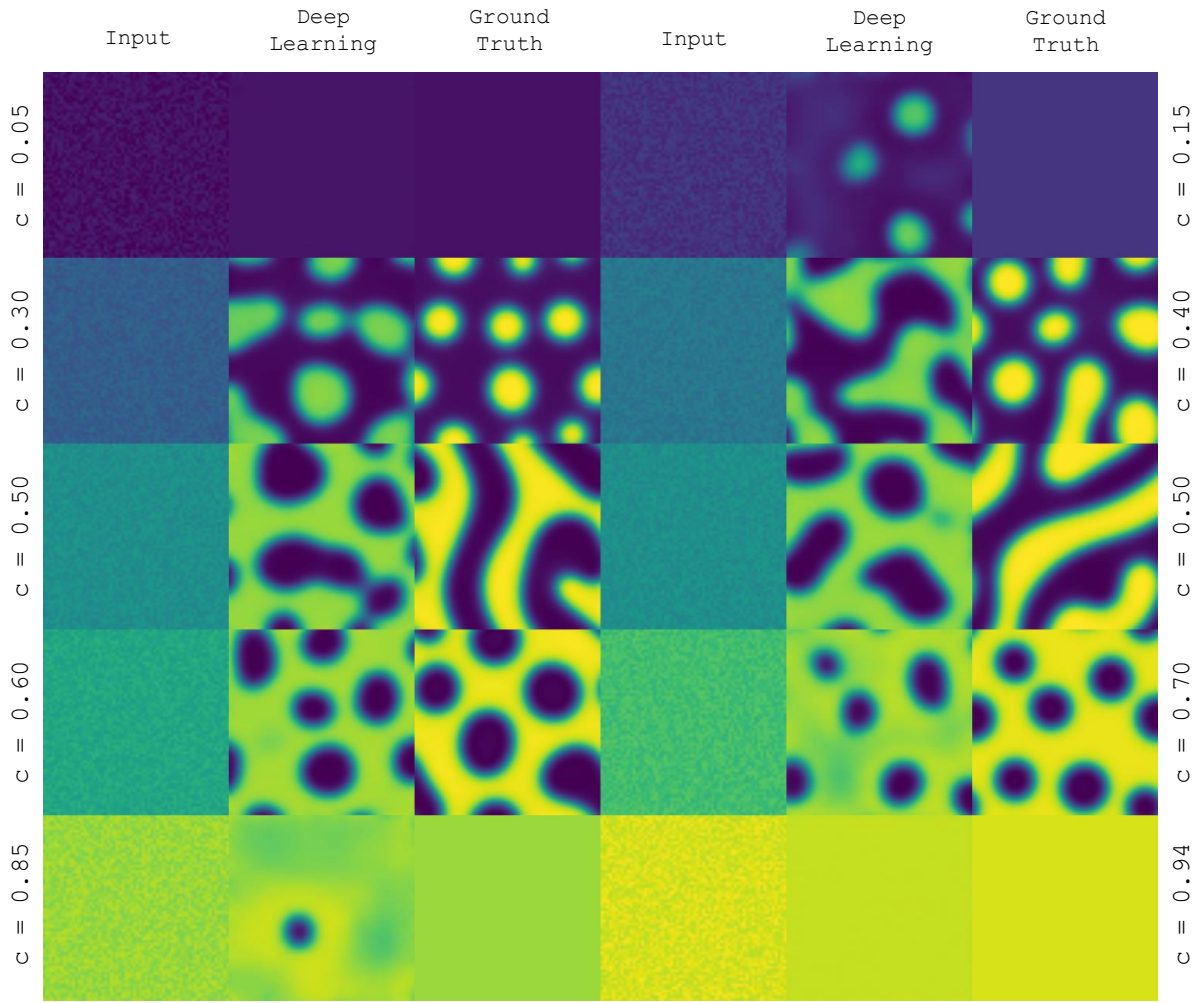

FIG. S8. **L1+100.0GAN performance on the phase segregation test set.** Long-time phase segregation behavior of a binary mixture predicted by deep learning. The initial input distribution of mixture (left), the predicted solution (center), the ground truth solution of the phase field (right) are shown for the concentration range c = 0.05-0.94. There are two samples represented for c = 0.50 to demonstrate the variety of outcomes.

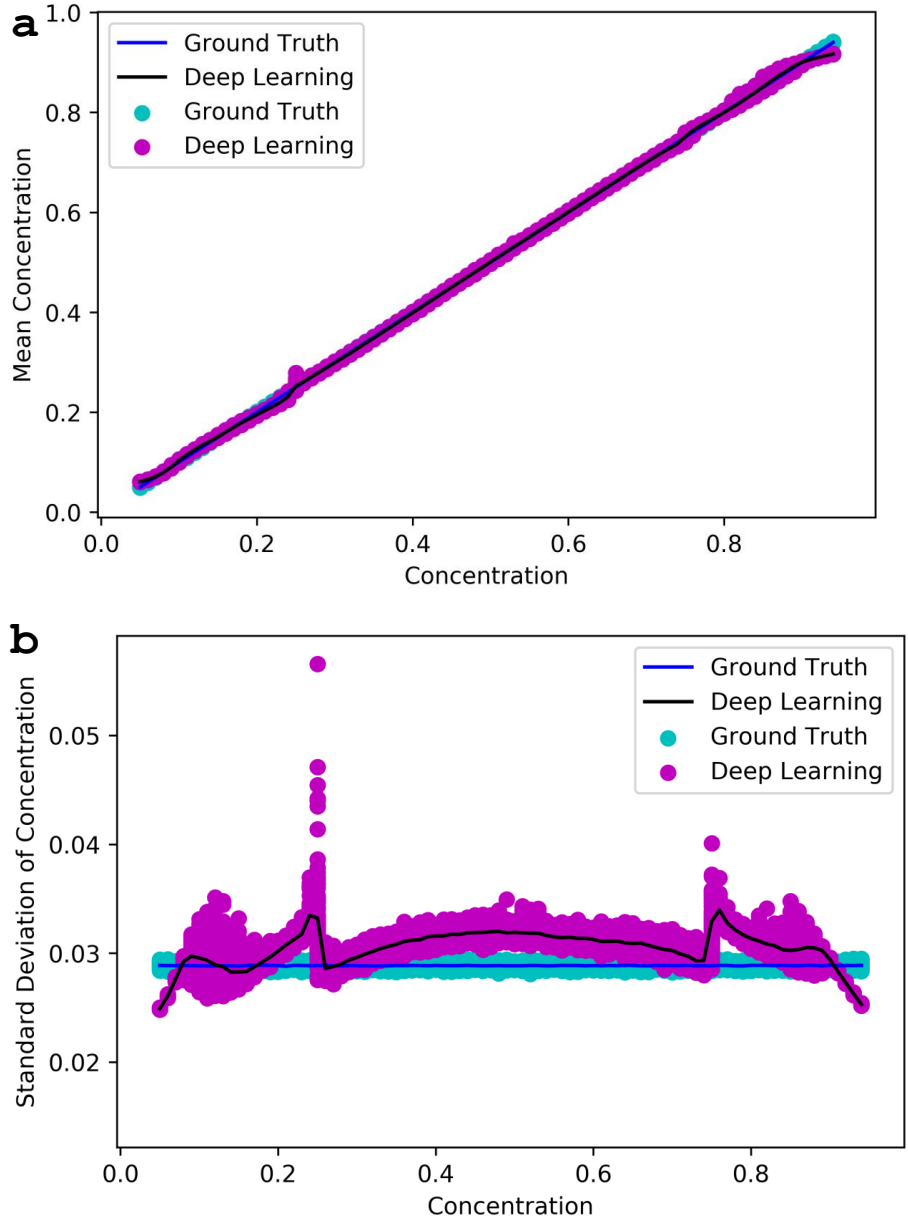

FIG. S9. **Comparison of metrics between deep learning model and ground truth for the L1+1.0GAN model on the reverse phase segregation test set. a)** Mean phase A concentration. **b)** Standard deviation of phase A concentration. The x-axis in all cases represents the initialized concentration of phase A.

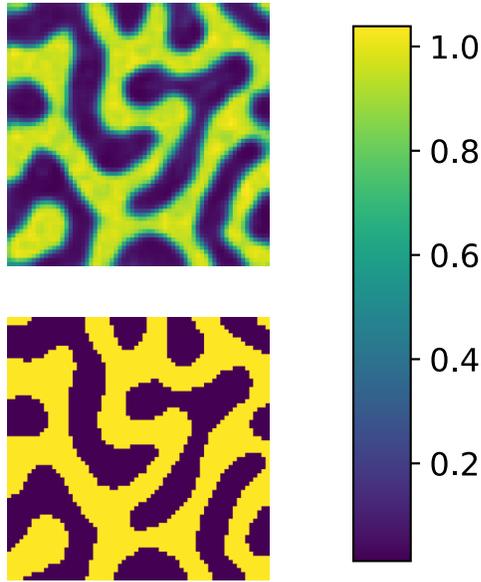

FIG. S10. *Original phase field (top) and binarized phase field (bottom) subject to a cutoff value of 0.5. Yellow represents phase A, while purple represents phase B.*

Subject to a cutoff that distinguishes phase A and phase B, the phase fields are binarized; concentrations that are higher than the cutoff are considered phase A (Figure S10). The binarized arrays are then used to calculate the overall size of the phase domains. In 2 dimensions, these size metrics are perimeter and area, both in units of pixels. To calculate the area, the number of non-zero elements is summed. To calculate the perimeter, we perform a Euclidean distance transform on the array and the number of terms with value 1 are summed; this number corresponds to the number of pixels in the A domain that directly contact the B domain. The analogs of these metrics in 3 dimensions are surface area and volume.

The Ginzburg-Landau free energy in the context of the Cahn-Hilliard problem represents the sum of the surface and chemical free energies over the whole domain as stated above.

The integration for each expression is performed using Simpson's rule.